%% file: main.tex
\documentclass{article}
\usepackage{spconf}
\usepackage{amsmath}
\usepackage{times}  
\usepackage{multirow}
\usepackage{helvet}  
\usepackage{courier}  
\usepackage[hyphens]{url}  
\usepackage{graphicx} 
\usepackage{nameref}
\usepackage{amssymb}
\usepackage{varioref}
\usepackage{algorithm}
\usepackage{algorithmic}
\usepackage{enumitem}
\setlist{nosep, leftmargin=14pt}

\usepackage{mwe} 

\title{Envisioning MedCLIP: A Deep Dive into Explainability for Medical Vision-Language Models}
\threeauthors
 {Anees Ur Rehman Hashmi$^1$, Dwarikanath Mahapatra$^2$,  Mohammad Yaqub$^1$}
	{$^1$Mohamed bin Zayed University of Artificial Intelligence, \\  $^2$Inception Institute of Artificial Intelligence\\
	Abu Dhabi, UAE}{}{}

\begin{document}
%
\maketitle
\begin{abstract}
Explaining Deep Learning models is becoming increasingly important in the face of daily emerging multimodal models, particularly in safety-critical domains like medical imaging. However, the lack of detailed investigations into the performance of explainability methods on these models is widening the gap between their development and safe deployment. In this work, we analyze the performance of various explainable AI methods on a vision-language model, MedCLIP, to demystify its inner workings. We also provide a simple methodology to overcome the shortcomings of these methods. Our work offers a different new perspective on the explainability of a recent well-known VLM in the medical domain and our assessment method is generalizable to other current and possible future VLMs.

\end{abstract}
\input{sections/introduction}

\input{sections/methodology}
\input{sections/results_discussion}
\bibliographystyle{IEEEbib}
\bibliography{strings,refs}

\end{document}

%% file: sections/introduction.tex
\section{Introduction}  \label{sec:intro}
\vspace{-0.5\baselineskip} 
Deep Learning (DL) models provide a remarkable performance on many tasks, however, they often work as a \textit{black box} and their internal working mechanism is hidden from the end-user \cite{von2021transparency}. This induces skepticism and a lack of trust in these models, particularly for cases like clinical diagnosis. Thereby hindering the real-life deployment and adaptation of DL models in critical domains like healthcare.

Explainable AI (XAI) \cite{arrieta2020explainable} offers a remedy to this problem by providing insights into the inner workings of DL models. It enables the end-users to see and understand the rationale behind models' prediction, enhancing fairness and confidence in DL models. In recent years, a number of XAI methods have been introduced to demystify these \textit{black-box} behavior in DL. These methods differ from each other mainly based on the information they offer, their applicability (model specific or agnostic), and the scope of their explanation \cite{arrieta2020explainable}. Furthermore, XAI methods can be categorized as intrinsic or post-hoc depending on the provided interpretation. Intrinsic or model-based methods rely on the structure of the model itself and hence are limited to models like linear regression. Whereas post-hoc methods are applied to more complex models that are often difficult to explain. Examples of post-hoc methods include gradient backpropagation \cite{simonyan2013deep}, layer-wise relevance propagation  \cite{bach2015pixel}, and class activation mapping (CAM) \cite{zhou2016learning}. 

Multimodal learning has shown great success in combining the information from various modalities to increase the performance of deep learning models \cite{baltrusaitis2018multimodal, bayoudh2021survey}. Particularly, vision-language models (VLM) utilize both visual and textual information to learn a meaningful representation \cite{girshick2014rich, jia2021scaling} and to answer complex questions related to the associated data \cite{antol2015vqa, zhang2023vision}. These models bring AI a step closer to the human-like working paradigm where AI models can process multiple types of inputs simultaneously to perform a task. It also enables the utilization of large-scale pre-trained vision and language models for downstream tasks, which reduces the cost and hustle of training models from scratch. 

VLMs have also demonstrated significant utility in medical applications, as evidenced by studies such as \cite{zhang2022contrastive, huang2021gloria, wang2022medclip}. Moreover, the medical domain contains the paired scans-reports that can be utilized to train VLMs for enhanced performance in downstream tasks like classification, segmentation, and image generation \cite{chambon2022roentgen, nasir2023multi, wang2022medclip}. A recent VLM, MedCLIP \cite{wang2022medclip} efficiently trained the CLIP-like \cite{radford2021learning} model using a modified objective for X-Ray image-report dataset. However, despite being very useful, the complexity of VLMs makes them less plausible causing obstruction to their wide applications. Explaining these VLMs is of high importance but also challenging as typical XAI methods work with one modality i.e. text or image alone \cite{vandervelden2022explainable}, which can be misleading as VLMs heavily rely on the interaction between text and image features. Furthermore, it is unclear whether single-modality-based explainability methods can explain VLMs. 

To address the problem of explainability in VLMs, we analyze the effectiveness of different existing XAI methods to explain a recently introduced VLM, MedCLIP \cite{wang2022medclip}. Furthermore, we propose a simple and effective method to overcome the shortcomings of these methods by combining the XAI methods with the text and image interaction in VLM. This increases the plausibility as well as provides a framework to understand what the VLM actually looks for when making a certain prediction.

%% file: sections/methodology.tex
\section{Methodology} 
\vspace{-0.5\baselineskip} 
We selected a diverse set of XAI methods for our analysis. A brief overview of these methods is as follows:
\begin{itemize}
    \item \textbf{Gradient backpropagation} (GB) is a saliency-map-based method that backpropagates the gradients to find the features of high importance \cite{springenberg2014striving}. It assigns each input feature an importance score which indicates that feature's contribution to the model prediction. 
    
    \item \textbf{Occlusion} method is used for analyzing the importance of each region in the input. It partly occludes an image and observes the change in the network activation and the model's prediction \cite{zeiler2014visualizing}. The input regions that bring the most change are considered to be the most important.
    
    \item \textbf{Integrated-Gradients} (IG) interpolates a baseline image to the actual input image to quantify the importance of each pixel. The baseline image can be random noise or pure white pixels. This is an efficient method that integrates the gradients at different levels \cite{sundararajan2017axiomatic}.
    
    \item \textbf{Grad-Shapley} (GS) combines the gradients of the model predictions with respect to the input pixels and combines them with the Shapley value \cite{selvaraju2017gradcam}. 
\end{itemize}

We selected these four XAI methods for several reasons. First, they represent a diverse range of techniques that offer distinct insights into the inner workings of the DL model. GB helps us understand feature importance through the gradient flow in the model. Occlusion provides insights into the model's sensitivity to different image regions by systematically occluding parts of the input. IG method offers a way to attribute predictions to specific features by tracking their influence across input variations. GS method, on the other hand, enables us to quantify the contribution of each feature's interaction to the final prediction. Moreover, these methods are well-established and widely recognized in the field of XAI. They have been applied successfully across various domains, making them suitable choices for our specific use case. 

For our analysis, we first applied these methods to MedCLIP in order to analyze their performance through explainability maps for a given image. This is followed by using our proposed method to acquire the explainability maps and analyze the difference in interpretability.



    
        
    

\begin{figure*}[h]
    \centering
    \includegraphics[width=0.92\textwidth, height=0.35\textheight]{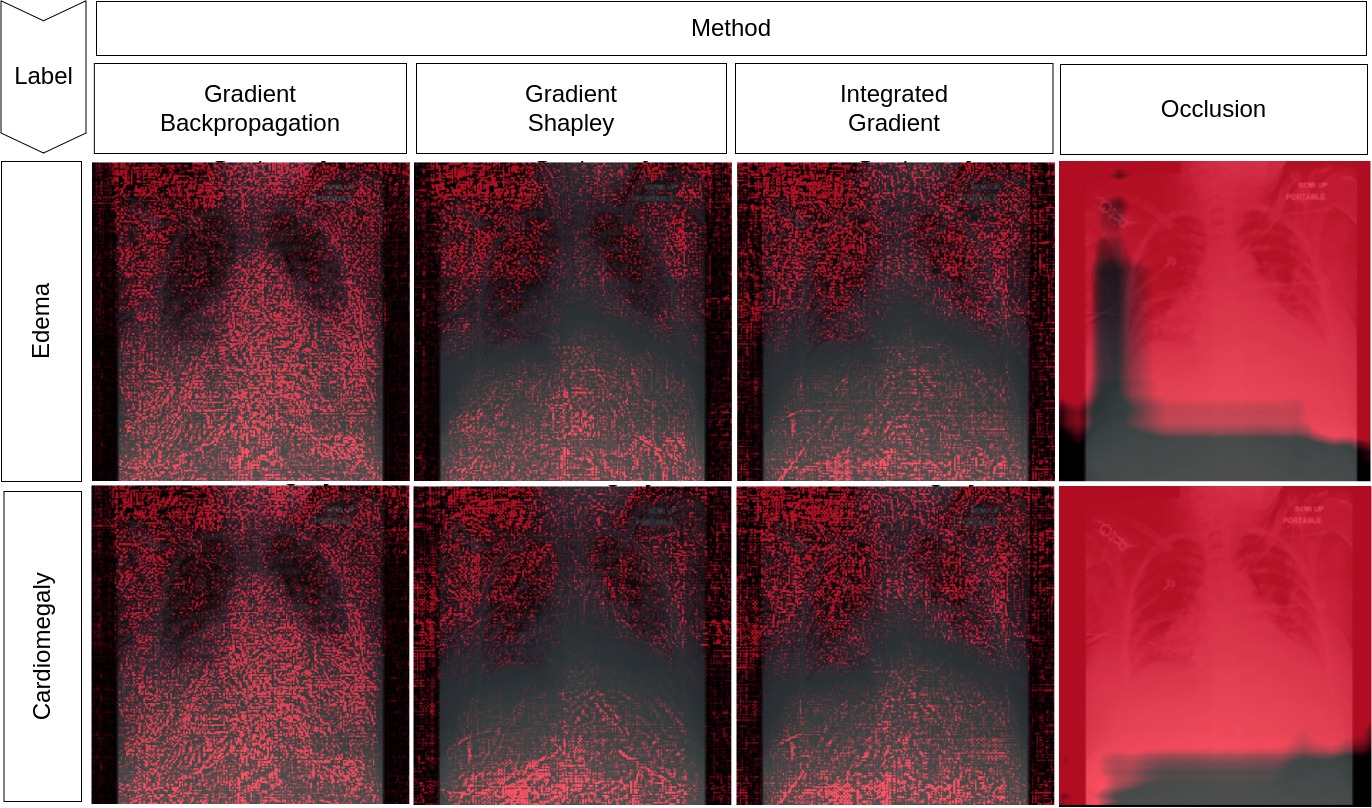}
    \caption{Class-specific feature maps generated for the prompt-classifier. These maps do not provide any significant class-specific information and therefore are not suitable for explaining VLM like MedCLIP.}
    \label{fig:clf_map}
\end{figure*}

\begin{figure*}[h]
    \centering    
    \includegraphics[width=0.92\textwidth, height=0.4\textheight]{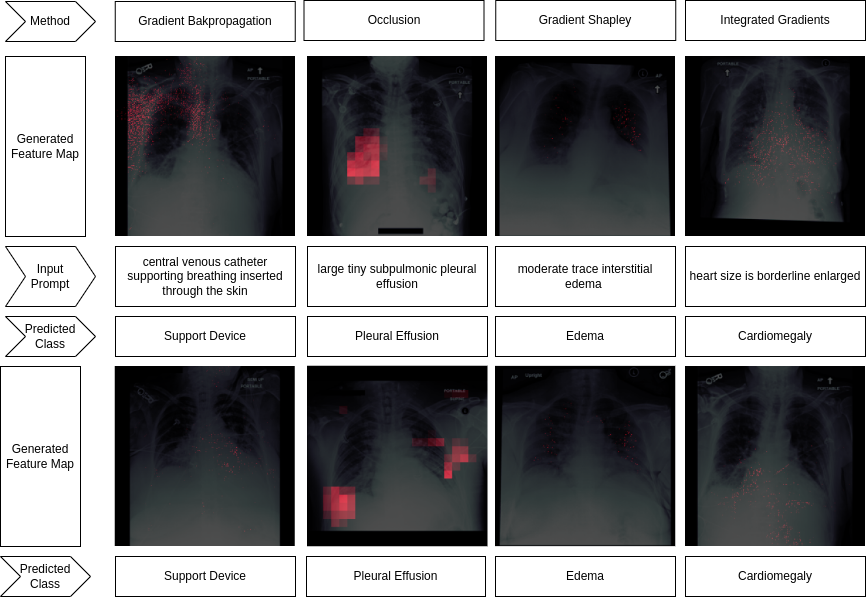}
    \caption{
            Feature activation maps were generated using the proposed methods. These activation maps are focused and clearly explain MedCLIP. The top part shows show the VLM is able to focus on specific areas of the input image based on the input prompts. Whereas, the lower part shows the activation maps for the used models when provided with the class labels as text inputs. These maps comprehensively explain the difference in the performance of MedCLIP based on the type of text input provided.
        }
    \label{fig:prompt-maps}
\end{figure*}
\vspace{-0.5\baselineskip} 

\subsection{MedCLIP} \label{sec:medclip_method}
 \vspace{-0.5\baselineskip} 
MedCLIP is a recently introduced powerful VLM for chest X-ray (CXR) classification. It combines a BioClinicalBERT\footnote{https://huggingface.co/emilyalsentzer/Bio\_ClinicalBERT}-based text-encoder backbone and a SwinTransformer \cite{liu2021swin} vision-encoder, pre-trained on the ImageNet dataset \cite{deng2009imagenet}. This BERT model has been pre-trained on the MIMIC-III dataset \cite{johnson2016mimic} containing the electronic health records from ICU patients. These transformer-based large models possess the ability to efficiently learn complex features from the input data. 

First, a 224x224 input image (X$_{img}$) is passed through the vision encoder (V$_{enc}$) to produce image embeddings ($I \in \mathbb{R}^D$). These embeddings are then further projected into a  lower-dimensional vector $I_p \in \mathbb{R}^M$ via a projection head denoted as $P_v$.%
\begin{align}
    I  &= V_{enc}(X_{img}) \\ 
    I_p & = P_v(I) 
    \label{eq:ip}
\end{align}
Similarly, the input text ($X_{txt}$) is tokenized and then encoded using the text encoder $T_{enc}$. The resultant vector ($T \in \mathbb{R}^E$) is subsequently projected to $T_p \in \mathbb{R}^M$ using the text projection head $P_t$, to match the vision and text embeddings dimensions denoted as M (M=512 in MedCLIP).
\begin{align}
    T &= T_{enc}(X_{txt}) \\ 
    T_p & = P_t(T)
    \label{eq:tp}
\end{align}
These text and image embeddings ($I_p$ and $T_p$) are normalized before calculating the dot product $M_{dot}$ in a contrastive manner.
\begin{align}
    M_{dot} &= I_p\cdot T_p \\
    L &= M_{dot} * \tau
 \label{eq:medclip_dot} 
\end{align}
Here $L$ represents the final output logit, indicating the similarity between the input image and the text and $\tau$ is the learnable temperature parameter. 
\vspace{-0.5\baselineskip} 
\subsection{Proposed Approach}
 \vspace{-0.5\baselineskip} 
Our proposed approach is based on the idea of applying the selected XAI methods to the embedding space of the VLM, instead of the final output of the model. This is motivated by the fact that VLMs process each input separately before fusing them to generate the final output (see \nameref{sec:medclip_method} section) and each of the encoders can be treated as separate models. Hence applying the existing XAI methods for each of the input modalities and then combining it with the other modality can lead to better explainability. The proposed method has three main steps:

\begin{enumerate}
    \item Firstly, an XAI method of choice $M_{xai}$ is applied to the image embeddings generated by the vision-encoder of $V_{enc}$ of the MedCLIP to generate an explainability map $F_{map}^i \in \mathbb{R}^{AxA}$ for each embedding dimension $i$. This yields M distinct maps, each highlighting the input image pixels important for one specific image embedding dimension.
    \vspace{-0.5\baselineskip} 
    \[F_{map}^i = M_{xai}(model=V_{enc}, target=i)\]
    where $i$ is the index of the image embedding $Ip \in \mathbb{R}^{1x512}$.
    
    \item In the second step, a text input $X_txt$ is selected and encoded through the text encoder $T_{end}$ to generate the embeddings $T_p \in \mathbb{R}^{1xM}$. These generated embeddings are scaled by the learned temperature parameter of the VLM.

    \item In the final step, a dot product between the image explainability maps generated in step 1 and text embeddings is calculated to get a $\mathbb{R}^{AxA}$ weighted average of these maps $F_{map}^{out}$. 
    This step calculates the similarity of each text embedding to the corresponding image embedding as well as quantifies their contribution to the final output, thereby providing a measure of how important each explainability map is for the models' prediction.
    \vspace{-0.5\baselineskip} 
    \[F_{map}^{out} =  T_p \cdot F_{map}^{all}\]
    where $F_{map}^{all} \in \mathbb{R}^{MxAxA}$ is a list containing $M$ generated feature maps.   
\end{enumerate}

This results in a single explainability map that highlights the specific image pixels influencing the model's confidence with regard to the given input. It is important to note that this method is very efficient as the second and third steps can be repeated for different prompt embeddings once the complete set of feature maps for an image is obtained, obviating the need for repeating step 1 for each input text. We experimented with both class labels as well as text prompts (sentences) as text input to the model. A set of 10 prompts was developed for each class label. Each generated text prompt encompasses information about the specific pathology, its location, intensity, and variations in sub-types (e.g. "mild linear \textit{atelectasis} at the mid lung zone"). This comparison between text prompts and class labels will help us visualize the effect of different input types on the VLM.

\vspace{-0.5\baselineskip} 
\subsubsection{Dataset} \label{sec:dataset}
\vspace{-0.5\baselineskip} 
We used the MIMIC-CXR \cite{johnson2019mimic} dataset which is a large CXR dataset with free-text radiology reports collected from the Beth Israel Deaconess Medical Center in Boston, MA. This dataset contains approximately 377,110 CXR-report pairs with 14 classes (13 pathologies and 1 healthy class). We incorporated a subset of 2000 randomly selected samples along with the class labels for our analysis. 

\subsubsection{Implementation Details}
\vspace{-0.5\baselineskip} 
We performed all experiments on a single Nvidia Quadro RTX 6000 GPU with 24GB of memory. The MedCLIP model was implemented using the PyTorch library \cite{paszke2019pytorch}, while we used the Captum library \cite{kokhlikyan2020captum} for off-the-shelf XAI methods. 

%% file: sections/results_discussion.tex
\vspace{-0.5\baselineskip} 
\section{Results}
\vspace{-0.5\baselineskip} 
Figure \ref{fig:clf_map} shows the class-wise explainability maps generated using the four selected XAI methods. It's worth noting that these explainability maps exhibit a remarkable degree of similarity despite having significantly different inputs. What's more, they collectively assign a substantial portion of the image as important, which results in a high rate of false positives. This consistent high false-positive behavior is observed across all four methods. In fact, these methods often highlight pixels outside the human body in chest X-rays as equally crucial for the final prediction. Furthermore, it is evident that the class labels do not exert a discernible influence on the final output. These results fail to align with the established medical diagnostic methodologies, which are typically lesion-specific, as different regions of a chest X-ray are critical for different diagnoses. This discrepancy further underscores the limitations of conventional XAI methods in effectively elucidating the mechanisms underlying the MedCLIP model.
Our proposed approach produces the feature maps depicted in Figure \ref{fig:prompt-maps}. We generated explainability maps using both text prompts (sentences) and class labels to investigate the influence of different text inputs. To begin, we applied our method to MedCLIP, utilizing images and text prompts describing various lesions. As depicted in the top section of Figure \ref{fig:prompt-maps}, our approach stands in contrast to conventional XAI methods by avoiding false positives and accurately highlighting the most important images. The highlighted pixel locations closely align with established clinical diagnostic procedures for the specified pathology. Additionally, our method effectively illustrates how MedCLIP's focus shifts based on the input text prompt, providing strong evidence of this VLMs' capacity to comprehend text and identify relevant image pixels. It is also evident this method is capable of delivering nuanced, comprehensive, and meaningful insights into the model's operation. Secondly, we employed class labels in conjunction with chest X-ray (CXR) images to evaluate our method, as shown in the bottom section of Figure \ref{fig:prompt-maps}. The highlighted image pixels are localized and exhibit variations across different class labels.
\vspace{-1\baselineskip} 
\section{Discussion and Conclusion}
\vspace{-0.5\baselineskip} 
In this paper, we analyzed the usefulness of existing XAI methods for VLMs and provided a simple and highly effective method to overcome their shortcomings for the multimodal models. Through our work, we demonstrate the efficacy of our proposed approach in comprehensively explaining the functioning of the MedCLIP VLM, a feat that the conventional approach fails to achieve. Moreover, our approach shows the combined effect of both input modalities on the VLM prediction which can be of high importance in VLMs. Furthermore, the explainability maps can help us understand the discrepancy in the performance of DL models. We visualize the difference in activation maps for two different text input forms in order to understand the effect of different types of text inputs. One major benefit of this work is the flexibility to use any off-the-shelf XAI method for VLMs, making the method versatile. Furthermore, it can also be adapted to other VLMs by following the image and text embedding fusion approach used in that specific model.

In conclusion, pretrained VLMs like MedCLIP have enormous potential to be used for downstream tasks without fine-tuning. However, it is very important to make AI more trustworthy by making these models explainable to the end user. Further research is required to design new VLM-specific XAI methods or frameworks for practical use of already existing methods.